RESEARCH ARTICLE

# Stochastic scheduling of autonomous mobile robots at hospitals

**Lulu Cheng[1], Ning Zhao[1]\*, Mengge Yuan[1], Kan Wu[2]**

1 Faculty of Science, Kunming University of Science and Technology, Kunming, Yunnan, China, 2 Business Analytics Research Center, Chang Gung University, Taoyuan City, Taiwan

\* zhaoning@kust.edu.cn

## Abstract

This paper studies the scheduling of autonomous mobile robots (AMRs) at hospitals where the stochastic travel times and service times of AMRs are affected by the surrounding environment. The routes of AMRs are planned to minimize the daily cost of the hospital (including the AMR fixed cost, penalty cost of violating the time window, and transportation cost). To efficiently generate high-quality solutions, some properties are identified and incorporated into an improved tabu search (I-TS) algorithm for problem-solving. Experimental evaluations demonstrate that the I-TS algorithm outperforms existing methods by producing high-quality solutions. Based on the characteristics of healthcare requests and the AMR working environment, scheduling AMRs reasonably can effectively provide medical services, improve the utilization of medical resources, and reduce hospital costs.

## 1 Introduction

In recent years, the aging population has become a prominent trend, which has subsequently increased the workload and pressure on medical staff. Autonomous mobile robots (AMRs) can alleviate the workload of medical staff, reduce their potential infection risks, and improve work efficiency by assisting in completing delivery tasks such as delivering medicines, meals, and documents [1]. Scheduling plays a crucial role in rational and effective planning, as it is regarded as an effective tool for achieving high utilization of resources [2]. Therefore, reasonable and effective planning of AMR service routes is necessary to ensure high-quality service.

The deterministic scheduling problem of mobile robots has garnered considerable attention in the literature. However, in practical scenarios, the travel speed of robots is subject to variations caused by road conditions, and the service time of each robot is uncertain due to diverse demands. This paper aims to investigate the scheduling problem of mobile robots operating in hospitals, where both travel and service times are subject to stochastic variations. Specifically, AMRs are employed for medicine delivery services, transporting medications from the warehouse (pharmacy) to designated points within the hospital (wards). Each delivery request must be completed within a specified time window.

This paper focuses on efficiently scheduling multiple AMRs in a stochastic environment. The primary objective is to minimize the hospital's daily cost, which includes AMR fixed costs, penalties for late arrivals, and transportation expenses. To solve this problem, various properties are identified and incorporated into an improved Tabu Search (I-TS) algorithm, which serves as a solution approach.

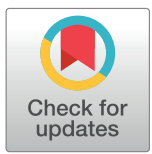

The aim of this research is to provide insights and practical strategies for optimizing the scheduling of AMRs in hospitals, taking into account the inherent uncertainties of travel and service times in a stochastic environment.

The organization of this paper is as follows. Section 2 is the literature review. A detailed description of the AMR scheduling problem is presented in Section 3. Section 4 presents the stochastic programming mathematical model. An improved tabu search (I-TS) algorithm is proposed in Section 5. Section 6 verifies the effectiveness of the algorithm with improved Solomon instances. The paper is concluded in Section 7.

## 2 Literature review

The scheduling problem of mobile robots has garnered significant attention in the literature. Most existing studies primarily focus on deterministic environments. Liu et al. [3] studied the vehicle routing problem with time windows and fixed running speed, where a robot may run multiple trips. They proposed an innovative two-index mixed integer programming model and obtained optimal solutions. Alitappeh and Jeddisaravi [4] studied multi-robot task allocation in deterministic environments to minimize the total system cost. The problem was partitioned into regions and routing sub-problems, which were solved by a multi-objective genetic algorithm and reinforcement learning approach. Jun et al. [5] also studied the AMR scheduling problem in the deterministic environment and introduced an innovative local search algorithm to solve it.

However, in practical scenarios, robots predominantly operate in stochastic environments. The scheduling problem of AMRs in a stochastic environment is a variant of the classic vehicle routing problem due to their service characteristics. The vehicle routing problem (VRP) typically involves three stochastic factors: (1) uncertainty in requests, including uncertain demands and random requests [6, 7, 8]; (2) uncertainty in service time [9–11]; and (3) uncertainty in travel time [12, 13]. Some researchers have studied different variants of VRP. For instance, Taş et al. [10] investigated the VRP with a soft time window and stochastic travel times, considering both cases with and without service times in the model. Pan et al. [14] focused on a multi-trip VRP with known service times, hard time windows, and variable running speeds. Similarly, Jie et al. [15] considered a time-dependent VRP that is affected by the traffic environment, while Dávila et al. [16] studied the VRP with hard time windows and stochastic service time.

In a stochastic environment, the VRP is typically modeled as a multi-stage stochastic programming model to minimize the expected total cost. There are two main types of stochastic programming models: chance-constrained programming (CCP) models and stochastic programming with recourse models [17]. The CCP model, proposed by Charnes and Cooper [18], is a powerful tool for modeling stochastic decision systems. It requires that the probability of the constraint condition is no less than a certain confidence level [17]. Zhang et al. [19] studied VRP with time windows and stochastic travel and service times. A stochastic programming model is established to ensure customer service levels. Ehmke et al. [20] analyzed the distribution of customers' start service time and arrival time with stochastic travel time, limiting the probability of violating the time window constraint by a chance constraint method. Li et al. [17] studied a stochastic VRP by formulating both chance-constrained programming and stochastic programming models with recourse. Shi et al. [21] solved the healthcare routing problem with stochastic travel and service times using a stochastic programming model with recourse.

The VRP is a well-known NP-hard problem. An exact algorithm cannot find the optimal solution within an acceptable time as the number of requests increases. Thus, heuristic algorithms are typically employed to obtain satisfactory solutions. The common heuristics algorithms include the genetic algorithm [22, 23], variable neighborhood search algorithm [24, 25], grey wolf optimizer [26], ant colony algorithm [27], gravitational search algorithm [28], and tabu search algorithm (TS) [29, 30]. The TS algorithm is an effective method for solving VRPs in stochastic environments. It can generate a solution close to the optimal one, but the running time is long [31]. Therefore, many scholars have proposed improvements on the TS algorithm. For instance, Taş et al. [10] combined the TS algorithm with an adaptive large neighborhood search algorithm to solve VRP with random travel times. Ge et al. [32] solved electric VRPs with stochastic demands by combining Clarke and Wright's saving method [33] with the TS algorithm. Li and Li [34] proposed an improved TS algorithm based on a greedy insertion algorithm to solve VRPs with stochastic travel and service times. Pan et al. [35] and Taş et al. [10] combined the adaptive large neighborhood search algorithm and TS algorithm to solve time-dependent VRPs with time windows.

This paper is devoted to studying an AMR scheduling problem in a stochastic environment at hospitals. The following constrains are considered. (1) The medical requests need to be served within a specified time window; (2) The AMR has a limited load capacity; (3) Each AMR may run multiple trips per day; (4) The AMR's travel time and the elevator running time are affected by the environment; (5) The service time of the medical request is stochastic. Note that the previous study on routing problems didn't comprehensively consider the above five factors. The above practical features cause the analysis of the AMR scheduling problem to be challenging. The classical VRP model isn't applicable for this problem. It is necessary to develop a suitable model and design a tailored algorithm to solve this problem.

## 3 Problem description

In this paper, the AMR scheduling problem in a stochastic environment is studied. The hospital needs to plan the number of AMRs involved in the daily service and the routes from the depot (pharmacy) to each request point. Since AMRs continuously work for 10 hours after being fully recharged [1], the AMR does not need to recharge during the one-day service.

Determining the optimal numbers of vehicles is the fundamental problem in the management of an AMR system [36]. Let $m$ be the total number of AMRs participating in the service, and it is the decision variable. Each AMR has an identical capacity $Q$ and fixed cost $\xi_1$ ($). Assume that $D$ is the depot set, $D = D_1 \cup D_2$, $D_1$ is the set of starting depots, and $D_2$ is the set of ending depots. The set $D_2$ is a dummy copy of the set $D_1$ with the same depot.

Assume that the AMRs can take elevators. Because AMRs need to avoid obstacles such as pedestrians, the AMR travel time per unit distance $V_r$ (s/m) and the elevator running time per unit floor $V_f$ (s/level) are random variables. Let $V_r$ (s/m). follow a normal distribution with mean $\mu(V_r)$ and variance $\sigma^2(V_r)$, and $V_f$ (s/level) follow a normal distribution with mean $\mu(V_f)$ and variance $\sigma^2(V_f)$. For the pedestrian flow at hospitals, we follow the approach proposed by Ichoua et al. [37] and divide the 24 hours a day into multiple non-overlapping time zones, which are 6:00–8:00, 8:00–12:00, 12:00–14:00, 14:00–16:00 and 16:00–24:00, where 8:00–12:00 is the morning peak, 14:00–16:00 is the afternoon peak, and the other time zones are the non-peak. $\mu(V_r)$ and $\mu(V_f)$ vary with the environment, as shown in Fig 1.

Let $R = \{1,2,\cdots\}$ denote the set of medical requests. For request $i$ ($i \in R$), the demand is $q_i$ ($q_i \leq Q$), the service time is $S_i$ and the time window is $[e_i, h_i]$, where the lower bound $e_i$ is hard

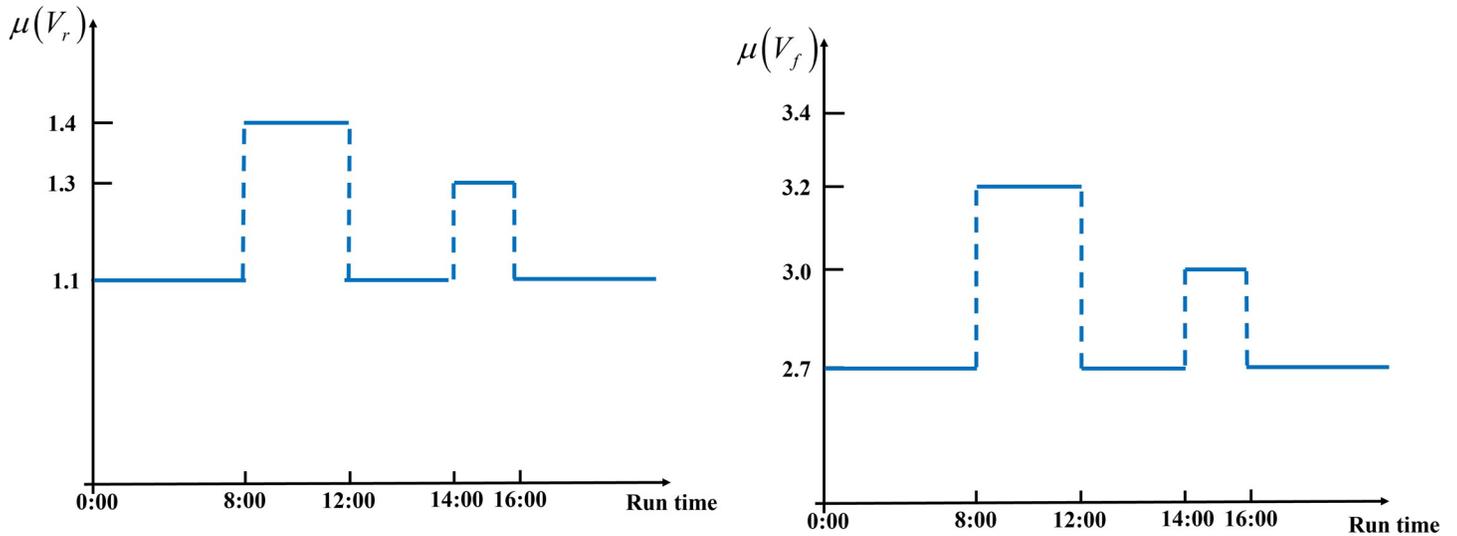

**Fig 1. Stepwise functions of $\mu(V_r)$ and $\mu(V_f)$.**

https://doi.org/10.1371/journal.pone.0292002.g001

(that is, the earliest service start time is $e_i$) and the upper bound $h_i$ is soft. If an AMR arrives later than $h_i$, the penalty cost for violating the time window is $\xi_2$. Let service time follow a normal distribution, i.e., $S_i \sim N(\mu(S_i), \sigma^2(S_i))$. The distance between medical requests $i$ and $j$ is $d_{ij}$, and their floor difference is $|f_{ij}|$, $i, j \in R$. Let $d_{ij} = d_{ji}$, $|f_{ij}| = |f_{ji}|$, $d_{ij} + d_{jk} \geq d_{ik}$. If $|f_{ij}| \neq 0$, we have $d_{ij} = d_{i,ele} + d_{ele,j}$, where $d_{i,ele}$ is the distance from request $i$ to the elevator, and $d_{ele,j}$ is the distance from the elevator to request $j$.

To improve the utilization of AMRs, each AMR can travel multiple trips. Let the route set traveled by the $k$th AMR be $L_k = \{1, 2, \ldots\}$, and the unit travel cost be $\xi_3$ (\$/s). The travel time spent by the $k$th AMR on the route $p \in L_k$ from request $i$ to request $j$ is $T_{pij}^k = \left(d_{ij} \cdot V_r + |f_{ij}| \cdot V_f\right) \cdot x_{pij}^k$, where $x_{pij}^k$ is a 0–1 variable and

$$x_{pij}^k = \begin{cases} 1, & \text{the } k\text{th AMR serves requests } i \text{ and } j \text{ consecutively on its } p\text{th route;} \\ 0, & \text{others.} \end{cases}$$

Because $V_r$ and $V_f$ follow normal distributions, we have $T_{pij}^k \sim N\left(\mu\left(T_{pij}^k\right), \sigma^2\left(T_{pij}^k\right)\right)$, where

$$\mu\left(T_{pij}^k\right) = \left(d_{ij} \cdot \mu(V_r) + |f_{ij}| \cdot \mu\left(V_f\right)\right) \cdot x_{pij}^k,$$

$$\sigma^2\left(T_{pij}^k\right) = \left(\left(d_{ij}\right)^2 \sigma^2(V_r) + |f_{ij}|^2 \cdot \sigma^2\left(V_f\right)\right) \cdot x_{pij}^k.$$

The $k$th AMR's remaining loading capacity is denoted by $u_{pi}^k$ when it arrives at request $i$ on its $p$th path. And $u_{pi}^k$ decreases $q_i$ after it completes the service of request $i$. The moment that it starts to serve request $i$ is denoted by $Y_{pi}^k = max\left\{A_{pi}^k, e_i\right\}$, where $A_{pi}^k$ is the time that the $k$th

AMR arrives at request $i$. If $x_{pij}^k = 1$, we have

$$A_{pj}^k = Y_{pi}^k + S_i + T_{pij}^k.$$

Fig 2 illustrates a simple example. It involves three AMRs, ten medical requests, and a depot (pharmacy). All AMRs depart from the depot and eventually return to the depot. The three elements in vector ($l_i$ level, $q_i$ kg, $[e_i, h_i]$) represent the floor, the demand, and the time window of the request point $i$, respectively.

Some notations for the AMR scheduling problem are defined in Table 1.

## 4 Model formulation

Since the travel time of AMRs between any two requests is affected by environmental factors such as traffic, congestion, and elevator availability, the AMR arrival time to a request is random. Under the condition that the travel time of the AMR, elevator waiting times, and service time of AMRs are stochastic, the AMRs are scheduled to minimize the daily cost of the hospital (including the AMR fixed cost, penalty cost of violating the time window and transportation

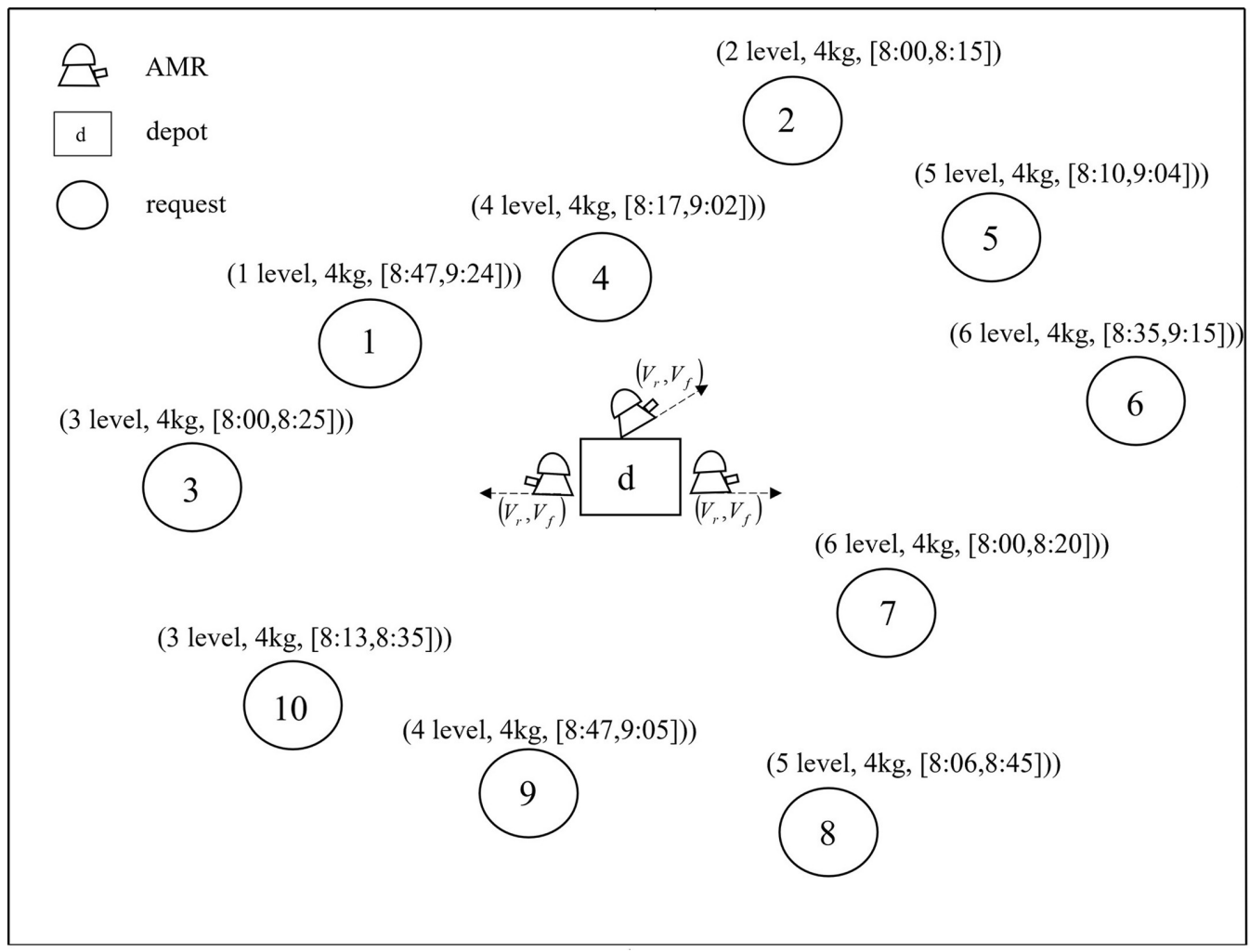

**Fig 2. An illustrated example.**

https://doi.org/10.1371/journal.pone.0292002.g002

Table 1. Notation for the AMR scheduling problem.

| Sets | |
|---|---|
| $D$ | Set of depots |
| $D_1$ | Set of starting depots |
| $D_2$ | Set of ending depots $D = D_1 = D_2 = \{d\}$) |
| $R$ | Set of medical requests |
| $L_k$ | Set of the route traveled by the $k$th AMR |
| **Parameters** | |
| $Q$ | Cargo capacity of the AMR |
| $\xi_1$ | Unit AMR cost |
| $\xi_2$ | The penalty cost for violating the time window |
| $\xi_3$ | Unit travel time cost |
| $q_i$ | Demand of request $i$ |
| $S_i$ | Service time of request $i$ |
| $[e_i, h_i]$ | Time window of request $i$ |
| $d_{ij}$ | Distance from request $i$ to request $j$ |
| $|f_{ij}|$ | Floor difference between requests $i$ and $j$ |
| $V_r$ | AMR travel time per unit distance |
| $V_f$ | Elevator running time per unit floor |
| $T_{pij}^k$ | The travel time spent by the $k$th AMR on the route $p \in L_k$ from request $i$ to request $j$ |
| $h_0$ | Any infinite large number |
| **Decision variables** | |
| $m$ | Total number of AMRs |
| $x_{pij}^k$ | 1 if the $k$th AMR travels arc $(i,j)$ on its $p$th route; 0 otherwise |

https://doi.org/10.1371/journal.pone.0292002.t001

cost). A stochastic programming model for the AMR scheduling problem is constructed as the following.

Objective function:

$$\min \ F(m, x) = \xi_1 \cdot m + \xi_2 \sum_{p \in L_k} \sum_{k=1}^{m} \sum_{i \in R} P\left(A_{pi}^k > h_i\right) \cdot x_{pij}^k + \xi_3 \sum_{p \in L_k} \sum_{k=1}^{m} \sum_{\substack{i \in D \cup R \\ j \in D \cup R \\ i \neq j}} \mu\left(T_{pij}^k\right) \cdot x_{pij}^k \quad (1)$$

s.t.

$$\sum_{k=1}^{m}\sum_{p\in L_k}\sum_{j\in R\cup D_2, i\neq j} x_{pij}^k = 1, \forall i \in R \quad (2)$$

$$\sum_{j\in R\cup D_2, j\neq i} x_{pij}^k = \sum_{j\in D_1\cup R, j\neq i} x_{pji}^k, \forall i \in D\cup R, k=1,2,\ldots,m, p\in L_k \quad (3)$$

$$q_i \leq u_{pi}^k \leq Q, \forall i \in D\cup R, k=1,2,\cdots,m, p\in L_k \quad (4)$$

$$q_j \leq u_{pj}^k \leq u_{pi}^k - x_{pij}^k \cdot q_i + Q(1-x_{pij}^k), \forall i \in D_1 \cup R, j \in R \cup D_2, i\neq j, k=1,2,\cdots,m, p \in L_k \quad (5)$$

$$Y_{pi}^k = \max\left\{A_{pi}^k, e_i\right\}, \forall i \in R, k=1,2,\cdots,m, p \in L_k \quad (6)$$

$$Y_{pi}^k = Y_{p+1,i}^k, \forall i \in D_1 \cap D_2, k=1,2,\cdots,m, p \in L_k \quad (7)$$

$$Y_{pi}^k + S_i + T_{pij}^k \cdot x_{pij}^k - h_0 \cdot \left(1-x_{pij}^k\right) \leq Y_{pj}^k, \quad \forall i \in D_1\cup R, j \in R\cup D_2, i\neq j, k=1,2,\cdots,m, p\in L_k \quad (8)$$

$$x_{pij}^k \in \{0,1\}, \quad \forall i \in D_1\cup R, j\in R\cup D_2, i\neq j, k=1,2,\ldots,m, p\in L_k \quad (9)$$

The objective function (1) is to minimize the total cost which is formed by three parts: the AMR fixed cost, the penalty cost of violating the time window, and transportation cost. The cost of one AMR is usually higher than the penalty cost of violating the time window. Without losing generality, let $\xi_1 > \xi_2 > \xi_3$. Constraint (2) ensures that each request is served in one and only one route. Constraint (3) indicates that the number of input arcs at each point is equal to the number of output arcs. The constraints (4)-(5) represent the remaining loading capacity constraints of the AMR on any route. The constraints (6)-(8) represent the time window constraints of the AMR arriving at every request on route $p$. The Eq (9) denotes a binary decision variable $x_{pij}^k$.

To more precisely represent the objective function (1), we take the approach of Ehmke et al. [20] to approximate the arrival time by a normal distribution, i.e., $A_{pi}^k \sim N\left(\mu\left(A_{pi}^k\right), \sigma^2\left(A_{pi}^k\right)\right)$. Let $x_{p,i,j}^k = 1$ $(i,j \in R\cup D)$, then $A_{pj}^k = Y_{pi}^k + S_i + T_{pij}^k$, where the starting service time $Y_{pi}^k = \max\left\{A_{pi}^k, e_i\right\}$. From Nadarajah and Kotz [38], we have

$$\mu\left(Y_{pi}^k\right) = \mu\left(A_{pi}^k\right)\Phi\left(\left(\mu\left(A_{pi}^k\right)-e_i\right)/\sigma\left(A_{pi}^k\right)\right) + e_i\Phi\left(\left(e_i - \mu\left(A_{pi}^k\right)\right)/\sigma\left(A_{pi}^k\right)\right) + \sigma\left(A_{pi}^k\right)\phi\left(\left(\mu\left(A_{pi}^k\right)-e_i\right)/\sigma\left(A_{pi}^k\right)\right) \quad (10)$$

$$\sigma^2\left(Y_{pi}^k\right) = \left(\mu^2\left(A_{pi}^k\right)+\sigma^2\left(A_{pi}^k\right)\right)\cdot\Phi\left(\left(\mu\left(A_{pi}^k\right)-e_i\right)/\sigma\left(A_{pi}^k\right)\right) + e_i^2\cdot\Phi\left(\left(e_i-\mu\left(A_{pi}^k\right)\right)/\sigma\left(A_{pi}^k\right)\right) - \left(\mu\left(Y_{pi}^k\right)\right)^2 + \left(\mu\left(A_{pi}^k\right)+e_i\right)\sigma\left(A_{pi}^k\right)\cdot\phi\left(\left(\mu\left(A_{pi}^k\right)-e_i\right)/\sigma\left(A_{pi}^k\right)\right), \quad (11)$$

where $\Phi(\cdot)$ is the cumulative distribution function of the standard normal distribution and $\phi(\cdot)$ is the probability density function of the standard normal distribution.

Since the travel time $T_{p,i,j}^k$ and the service time $S_i$ are independent random variables, we have $\mu\left(A_{p,j}^k\right) = \mu\left(Y_{pi}^k\right) + \mu(S_i) + \mu\left(T_{p,i,j}^k\right)$, and $\sigma^2\left(A_{p,j}^k\right) = \sigma^2\left(Y_{pi}^k\right) + \sigma^2(S_i) + \sigma^2\left(T_{p,i,j}^k\right)$. The probability of violating the time window $P\left\{A_{p,j}^k > h_j\right\}$ can be determined through $\mu\left(A_{p,j}^k\right)$ and $\sigma^2\left(A_{p,j}^k\right)$.

## 5 Improved tabu search algorithm

Due to computational complexity of the AMR scheduling problem, it is difficult to obtain an optimal solution within limited computation time. Hence, a heuristic algorithm is typically employed to find suboptimal solutions. Tabu Search (TS) algorithm is a heuristic algorithm based on neighborhood search proposed by Glover [39]. It can solve a class of combinatorial optimization problems including variants of the VRP that involve stochastic factors. The TS algorithm starts from an initial feasible solution and prevents previously visited solutions from being accepted again by setting a tabu list and a tabu tenure during the iterative search. In this section, we propose an improved tabu search (I-TS) heuristic algorithm to improve the quality of the solution and the computation speed. The I-TS algorithm improves the standard TS algorithm by incorporating a greedy insertion algorithm, adopting an adaptive mechanism for selecting neighborhood operators, and setting more effective tabu tenure parameters. These enhance the performance and effectiveness of the algorithm in solving the scheduling problem at hand.

The structure of Section 5 is outlined as follows: Section 5.1 covers the creation of an initial solution. In Section 5.2, we introduce three enhanced operators along with a repair operator. Section 5.3 delves into the tabu list and tabu tenure. Section 5.4 presents the aspiration criterion within the algorithm. Lastly, Section 5.5 introduces the termination condition for the I-TS algorithm.

### 5.1 Initial solutions

The I-TS algorithm performs an iterative search process to improve the solution by iteratively exploring the neighborhood of the current solution. The initial solution is crucial because it determines the starting point for the iterative search and affects the speed and quality of the algorithm's convergence to a good solution. To find a relative optimal solution more efficiently, the greedy insertion algorithm is adopted to generate a feasible initial solution. Algorithm 1 provides the pseudocode for the greedy insertion algorithm used to generate the initial solution for the I-TS algorithm. Requests are inserted sequentially into the current set of routes in ascending order by their time window, subject to capacity constraints, until all requests are processed.

**Algorithm 1. Greedy Insertion Algorithm for a Feasible Solution**

**Step 1** Arrange the lower bound $e_i$ of the time window of all requests in ascending order to generate a sequence of service priorities for requests $P = \{p_1, p_2, \ldots, p_i, \ldots, p_n\}$. Let $i = 1$ and the current route set is empty ($R_{current} = \Phi$).

**Step 2** Assign $p_1$ to a new AMR, and generate sub-route $r_1 = \{0 \to p_1\}$. $R_{current} \leftarrow R_{current} \cup r_1$.

**Step 3** If $i \leq n$

**then** repeat **Step 4**

**Else**

**then** go to **Step 5**.

**Step 4** If there exists a sub-route $r_j \in R_{current}$, $j = 1,2,...$ such that the request $p_i$ can be served at the end of $r_j$, and the constraints of the time window and capacity are satisfied..

**then** update sub-route $r_j = \{r_j \rightarrow p_i\}$, set $i \leftarrow i + 1$.

**Else**

**then** assign $p_i$ to a new AMR and generate a new sub-route $r_{new} = \{0 \rightarrow p_i\}$. Set $i \leftarrow i + 1$, $R_{current} \leftarrow R_{current} \cup r_{new}$.

**End**

**Step 5** Let $R_{current}$ be the initial feasible solution $x_0$, and output the corresponding objective function $F(x_0)$.

### 5.2 Neighborhood structure

This study uses three operators to generate neighborhood solutions. It is worth noting that in the process of using operators to generate neighborhood solutions, the elements used for transformation are usually randomly selected. To minimize the generation of infeasible neighboring solutions, particularly those violating time window constraints, we propose Proposition 1 and Proposition 2.

**Proposition 1.** Let the time windows of two requests $i_1$ and $i_2$ be $\left[e_{i_1}, h_{i_1}\right]$ and $\left[e_{i_2}, h_{i_2}\right]$, respectively, and $e_{i_1} < h_{i_1} < e_{i_2} < h_{i_2}$. If request $i_1$ is served before $i_1$ by the $k$th AMR on its $p$th route, then we have $P\left\{A_{p,i_1}^k > h_{i_1}\right\} = 1$ and $P\left\{Y_{p,i_1}^k > h_{i_1}\right\} = 1$.

**Proof.** Because request $i_2$ is served before $i_1$ by the $k$th AMR on its $p$th route, we have $Y_{p,i_1}^k \geq A_{p,i_1}^k > Y_{p,i_2}^k$. Since $Y_{p,i_2}^k = \max\left\{A_{p,i_2}^k, e_{i_2}\right\} \geq e_{i_2} > h_{i_1}$, it follows that $P\left\{A_{p,i_1}^k > h_{i_1}\right\} = P\left\{A_{p,i_1}^k > Y_{p,i_2}^k > h_{i_1}\right\} = 1$ and $P\left\{Y_{p,i_1}^k > h_{i_1}\right\} = P\left\{Y_{p,i_1}^k > Y_{p,i_2}^k > h_{i_1}\right\} = 1$.

**Remark:** From Proposition 1, we find that if the time windows satisfy $e_{i_1} < h_{i_1} < e_{i_2} < h_{i_2}$, serving request $i_1$ first is preferred, otherwise the time window of request $i_1$ is broken definitely.

**Proposition 2.** Let the time windows of two requests $i_1$ and $i_2$ be $\left[e_{i_1}, h_{i_1}\right]$ and $\left[e_{i_2}, h_{i_2}\right]$, respectively, and $h_1 < h_2$. The arrival times of the two requests are $A_{p,i_1}^k$ and $A_{p,i_2}^k$, which follow the same distribution. Then we have $P\left\{Y_{p,i_1}^k > h_{i_1}\right\} > P\left\{Y_{p,i_2}^k > h_{i_2}\right\}$.

**Proof.** The probability that request $i_1$ breaks its time window is

$$P\left\{Y^k_{p,i_1} > h_{i_1}\right\} = P\left\{\max\left(A^k_{p,i_1}, e_{i_1}\right) > h_{i_1}\right\} = 1 - P\left\{A^k_{p,i_1} \leq h_{i_1}\right\}P\left\{e_{i_1} \leq h_{i_1}\right\}$$
$$= 1 - P\left\{A^k_{p,i_1} \leq h_{i_1}\right\}.$$

The probability that request $i_2$ breaks its time window is

$$P\left\{Y^k_{p,i_2} > h_{i_2}\right\} = P\left\{\max\left(A^k_{p,i_2}, e_{i_2}\right) > h_{i_2}\right\} = 1 - P\left\{A^k_{p,i_2} \leq h_{i_2}\right\}P\left\{e_{i_2} \leq h_{i_2}\right\}$$
$$= 1 - P\left\{A^k_{p,i_2} \leq h_{i_2}\right\}.$$

Note that $A^k_{p,i_1}$ and $A^k_{p,i_2}$ follow the same distribution. Let the density of $A^k_{p,i_1}$ and $A^k_{p,i_2}$ be $p(x)$, then

$$P\left\{A^k_{p,i_1} \leq h_{i_1}\right\} = \int_0^{h_{i_1}} p(x)dx < \int_0^{h_{i_2}} p(x)dx = P\left\{A^k_{p,i_2} \leq h_{i_2}\right\}.$$

Therefore, $P\left\{Y^k_{p,i_1} > h_{i_1}\right\} > P\left\{Y^k_{p,i_2} > h_{i_2}\right\}$.

**Remark:** From Proposition 2 we find that if the time windows $\left[e_{i_1}, h_{i_1}\right]$ and $\left[e_{i_2}, h_{i_2}\right]$ satisfy $h_{i_1} < h_{i_2}$, prioritizing the service of request $i_1$ leads to a smaller probability of time window violation.

To obtain improved solutions within a limited time, we enhance three operators, namely swap*, 2-opt*, and relocation*, based on Proposition 1 and Proposition 2.

1. swap* operator: Fig 3(A) illustrates a scenario where request 5 violates its time window constraints at its current position, while the upper limit of request 2's time window is more relaxed. Consequently, requests 2 and 5 are chosen for swapping. On the other hand, if all requests on the current route satisfy their time window constraints, to avoid getting stuck in local optima, two requests from the current solution are randomly selected for position swapping, while the remaining requests remained unchanged. Fig 3(B) displays an example where requests 3 and 6 are selected, resulting in a new solution generated through the swap operation.

2. 2-opt* operator: As shown in Fig 4(A), a declining trend is observed in the lower bound of time windows for the sub-route between request 2 and request 5 in the existing route. Therefore, we reverse the order of this sub-route between requests 2 and 5 to improve the solution. For cases where there is no sub-route with decreasing time window lower bounds in the current solution, we randomly select two requests from the incumbent solution and reverse the order of points between them before inserting them back into the service route. As shown in Fig 4(B), requests 3 and 6 are randomly selected, and their routes are reversed to create a new feasible solution.

3. relocation* operator: By examining the time windows of requests on the current route, as shown in Fig 5(A), we notice that request 5 violates its time window constraints. Therefore, we select request 5 and remove it from its current position, then insert it before request 3 which has a later time window. If none of the above conditions apply to the current route, we randomly select a request, remove it from its current position, and insert it at a random position. As shown in Fig 5(B), request 6 is randomly selected and placed after request 2 in the new solution.

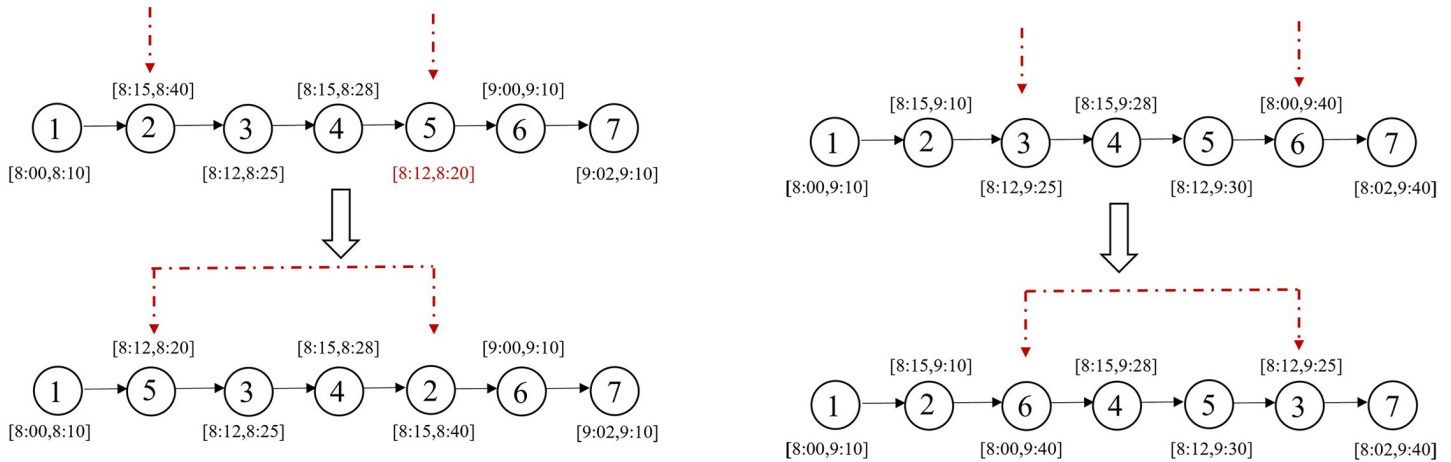

**Fig 3. Swap\* operator.**

https://doi.org/10.1371/journal.pone.0292002.g003

4. repair operator: In the metaheuristic algorithm, the search space may include infeasible solutions or tend towards suboptimal outcomes due to constraint violations [40]. Therefore, the solutions generated by the neighborhood operators previously mentioned (swap\*, 2-opt\*, and relocation\*) might be infeasible because they do not take into account load constraints. We introduce a repair operator, namely depot insertion. If a service route is not feasible due to overloading, a depot is inserted before the point where the remaining load does not meet the downward load demands, as shown in Fig 6. By implementing this repair mechanism, the infeasible solutions caused by constraint violations can be effectively repaired.

To find a high-quality feasible solution that satisfies all problem constraints as quickly as possible, the selection of the operators follows a roulette wheel procedure, where the probability $P_i = \frac{\rho_i}{\sum_{j=1}^{3} \rho_j}$ of selecting operator $i$ depends on its weight $\rho_i$, $i = 1,2,3$. The greater the weight

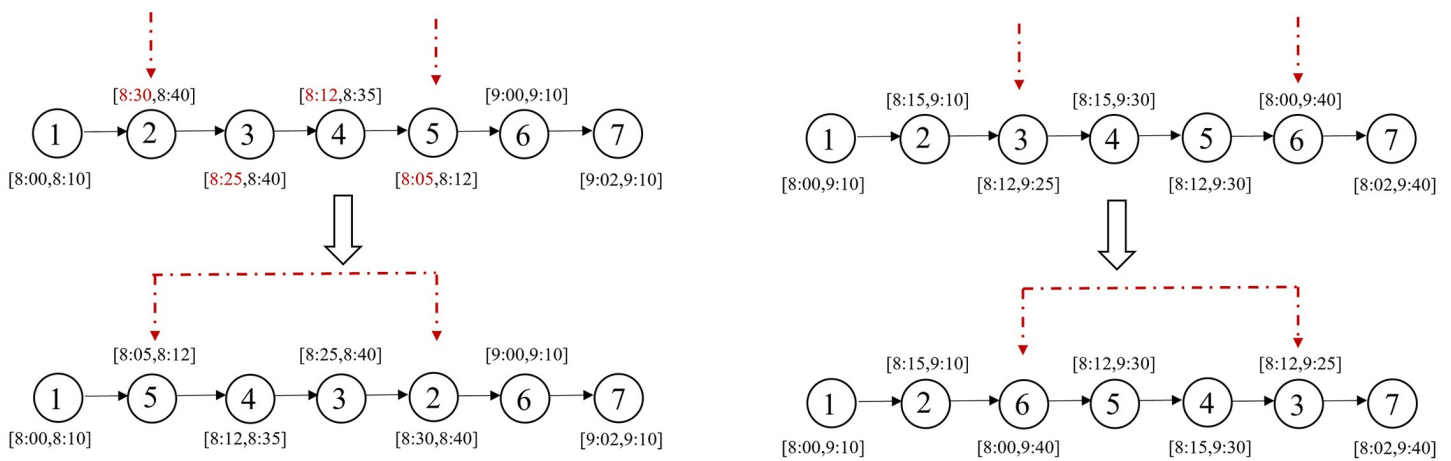

**Fig 4. 2-opt\* operator.**

https://doi.org/10.1371/journal.pone.0292002.g004

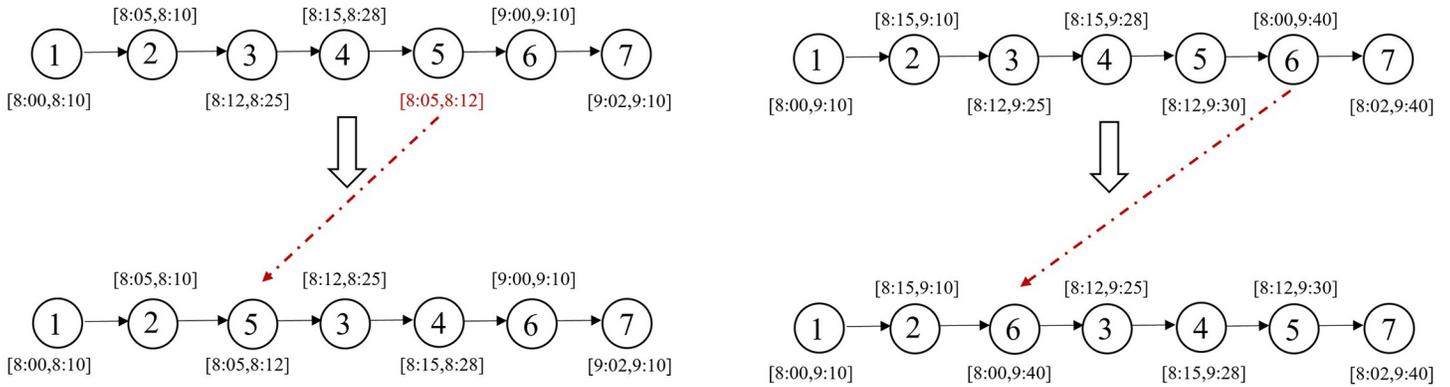

**Fig 5. Relocation* operator.**

https://doi.org/10.1371/journal.pone.0292002.g005

of operator $i$, the greater the probability of operator $i$ being selected. The weights $\rho_i$ are dynamic updated according to a reward scheme. If the neighborhood solution generated by operator $i$ is better than the current optimal solution, then $\rho_i = \rho_i + \delta_1$; if the neighborhood solution generated by operator $i$ is not tabu but produces a solution worse than the current optimal solution, then $\rho_i = \rho_i + \delta_2$, $\delta_1 > \delta_2 \geq 0$. The three operators in the initial state have equal weights and let $\rho_i = 1$, i.e., the probability of operator $i$ being selected is $P_i = \frac{1}{3}$. To find high-quality solutions more efficiently, the algorithm updates the weights $\rho_i$ of the operators after every ten iterations.

### 5.3 Tabu list and tabu tenure

The tabu mechanism in the I-TS algorithm can explore a wide range of solutions and avoid getting stuck in local optima in the search process. Here we use three tabu lists $B_i$, $\forall i = 1,2,3$ to store the solutions produced by applying each of the three neighborhood operators (swap*,

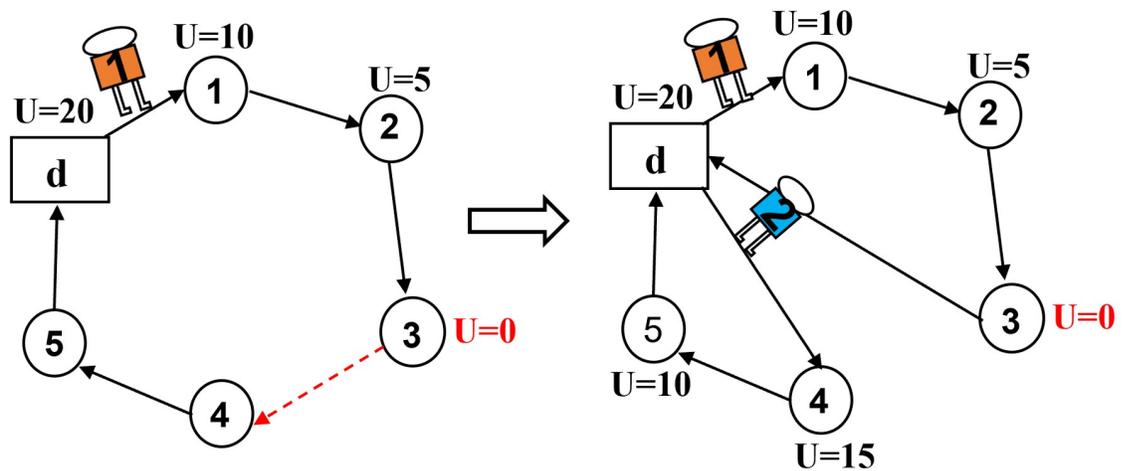

Note: U is the remaining capacity after serving the request

**Fig 6. Depot insertion operation.**

https://doi.org/10.1371/journal.pone.0292002.g006

2-opt*, and relocation*) to each pair of requests, where the tabu list $B_i$ is a $n \times n$ symmetric matrix, and $n$ is the total number of requests. For the operator $i$, the element $B_i(j_1, j_2)$ in the tabu list $B_i$ represents the tabu state of the neighborhood action $b_i(j_1, j_2)$, where $j_1, j_2 \in R$ are the two requests for swap, inversion or relocation.

For the tabu tenure in the I-TS algorithm, a long tabu tenure may cause the algorithm to overlook better solutions, while a short tabu tenure may lead to premature convergence to suboptimal solutions. Therefore, we consider choosing an appropriate tabu tenure value $t$ that balances exploration and exploitation (See Section 6.2 for details).

### 5.4 Aspiration criterion

To prevent the algorithm from prematurely terminating in a suboptimal solution, an aspiration criterion is adopted. For a neighborhood solution that is currently tabu but better than the current optimal solution, we revoke the tabu operation and take the solution as the current optimal solution.

### 5.5 Algorithm stop condition

The algorithm stops after a predetermined maximum number of iterations $N$ and outputs the current optimal solution.

Algorithm 2 provides the pseudocode for the I-TS algorithm.

**Algorithm 2. Improved Tabu Search (I-TS) for the AMR Scheduling Problem**

**Step 1** Initialize the weights of the three operators $\rho_i = 1, i = 1, 2, 3$ and $\delta_1, \delta_2$, three tabu lists $B_i, i = 1, 2, 3$, tabu tenure $t$ and the algorithm iterations $ite = 1$.

**Step 2** Let $x_0$ (obtained from **Algorithm 1**) be the initial solution of the algorithm, i.e., $x_{current} = x_0$, $F(x_{current}) = F(x_0)$. And the current best solution is $x_{best} = x_0$, $F(x_{best}) = F(x_0)$.

**Step 3** Based on the current solution $x_{current}$, use the roulette wheel procedure for the three operators (swap*, 2-opt*, and relocation*) to select the operator $i$ and generate the corresponding neighborhood $N(x_{current})$. The depot insertion operator is used to repair neighborhood solutions. Find the optimal objective value $F(x_{c\_best})$ and the corresponding neighborhood action $b_i(j_1^*, j_2^*)$, where $x_{c\_best} \in N(x_{current})$.

**Step 4 if** $F(x_{c\_best}) < F(x_{best})$

 **then** $F(x_{best}) = F(x_{c\_best})$, $x_{best} = x_{c\_best}$, $x_{current} = x_{c\_best}$, $F(x_{current}) = F(x_{c\_best})$, the weight $\rho_i$ of the selected operator $i$ is updated to $\rho_i = \rho_i + \delta_1$ through the reward scheme, and update the tabu list. Let $B_i(j_1^*, j_2^*) = \begin{cases} 0, & \text{if } B_i(j_1^*, j_2^*) \neq 0 \\ t, & \text{if } B_i(j_1^*, j_2^*) = 0 \end{cases}$, $B_i(j_1, j_2) = B_i(j_1, j_2) - 1$ if $B_i(j_1, j_2) \neq 0$ and $j_1 \neq j_1^*, j_2 \neq j_2^*$.

 **Elseif** $F(x_{c\_best}) \geq F(x_{best})$

> **then** find the optimal solution $x'_{c\_best}$ generated by the neighborhood action $b_i(j'_1, j'_2)$ for $\forall B_i(j'_1, j'_2) = 0$. Let $x_{current} = x'_{c\_best}$, $F(x_{current}) = F(x'_{c\_best})$, the weight $\rho_i$ of the selected operator $i$ is updated to $\rho_i = \rho_i + \delta_2$ through the reward scheme, and update the tabu list. Let $B_i(j'_1, j'_2) = t$, $B_i(j_1, j_2) = B_i(j_1, j_2) - 1$ if $B_i(j_1, j_2) \neq 0$.
>
> **Step 5** Let $ite = ite + 1$. The probability of each operator $P_i = \dfrac{\rho_i}{\sum_{j=1}^{3} \rho_j}$ is updated after every 10 iterations. Repeat **Steps 3–4**. When $ite = N$, stop the algorithm and output the current optimal solution $F(x_{best})$ and the corresponding set of routes $x_{best}$.

## 6 Numerical experiments

This section verifies the performance of the proposed I-TS algorithm. Section 6.1 introduces the improved Solomon instances. The impact of different tabu tenure $t$ and the number of algorithm iterations $N$ on the objective value is analyzed in Section 6.2. In Section 6.3, we compare the results of I-TS and CPLEX in a deterministic environment. For the stochastic scheduling of AMRs, we compare I-TS with VNS, TS, and greedy insertion algorithms. An AMR routing planning example is presented in Section 6.4. Section 6.5 presents numerical examples that provide managerial insights into the participation of AMRs in medical services.

### 6.1 Improved Solomon instances

As there is no suitable benchmark set of problems available for the AMRs service, we generate new instances based on three instances in the Solomon benchmark (C, R, and RC). The C instance has a clustered distribution of geographical locations for the requests, while the R instance has a uniform distribution, and the RC instance is relatively semi-clustered. Each instance comprises two subsets: type-1 and type-2. The time window of type-1 instances (C1, R1, and RC1) is narrow, whereas that of type-2 instances (C2, R2, and RC2) is relatively wide.

Note that $\mu(V_r)$ and $\mu(V_f)$ are dependent on the environment. We consider three distinct environments at hospitals: morning peak (P1), afternoon peak (P2), and off-peak (P3). For each environment, six instances (C108, C208, R101, R202, RC101, and RC202) are analyzed. Each instance comprises 20, 50, or 100 requests, where the first 20 and first 50 requests are selected from the total 100 requests. For example, P1-C108-100 indicates the instance C108 with 100 requests during the morning peak. In total, 54 instances are generated based on geographical location distribution, time windows, environments, and the number of requests.

The Solomon benchmark has been modified by making the following adjustments: (1) The earliest service time of the pharmacy is 7:30 AM (with an initial time of 0), and the latest service time is 7:30 AM in the following day. (2) Based on hospital traffic in one day, 8:00–12:00 and 14:00–16:00 are designated as the morning and afternoon peak periods, respectively, while the rest of the time is considered non-peak. (3) The AMR travel and elevator running speeds are assigned as follows: $\mu(V_r) = 1.4s/m$ and $\mu(V_f) = 3.2s/level$ during the morning peak period, $\mu(V_r) = 1.3s/m$ and $\mu(V_f) = 3.0s/level$ during the afternoon peak period, and $\mu(V_r) = 1.1s/m$ and $\mu$ ($V_f$) = 2.7$s$/$level$ during non-peak period. $\sigma^2$ ($V_r$) and $\sigma^2$ ($V_f$) are the same among the three periods. (4) The medical requests are distributed across floors 1 – 6, and the elevator is located at the center of each floor. In the Solomon instance, $\mu$ ($S_i$) represents the requested service time, and its variance is represented by $\sigma^2$ ($S_i$) = 15.

All of the following experiments were implemented in Matlab 2018b and were run on a microcomputer with a 3.60GHz CPU and 16GB of RAM. The environment used an Inter(R) Core(TM) i9-9900K CPU@3.60GHz, 16.00GB and Windows 10 operating system. Table 2 shows the parameter settings in the experiments.

## 6.2 Parameter tuning

Parameter tuning is crucial to ensure effective performance of the I-TS algorithm. This subsection presents experiments on two key parameters, tabu tenure $t$ and number of iterations $N$, using examples with 50 and 100 requests during non-peak periods (P3) in hospitals. Results are shown in Figs 7 and 8. Fig 7. depicts the algorithm's search for optimal solutions with tabu tenures of $t$ = 20, 30, 40, 50. As in Fig 7, the algorithm converges faster and achieves relatively optimal objective values when the tabu tenure is $t$ = 40.

Fig 8 presents the fitness curve of total hospital cost generated by the I-TS algorithm for non-peak cases. As shown in Fig 8, the objective value gradually approaches the optimal value as the number of iterations increases and the objective value converges after around 400 iterations. To balance algorithm effectiveness and runtime, let the iteration number $N$ equal to 500 for the I-TS algorithm.

Based on the experiments described above, let the tabu tenure $t$ = 40 and the maximum number of iterations $N$ = 500 in the following subsections to effectively solve the AMR scheduling problems.

## 6.3 Validity test of I-TS algorithm

This section evaluates the effectiveness of the proposed model and algorithm. Since the CPLEX optimizer is limited to small instances, our algorithm is first validated by solving 20-request instances in a deterministic setting using CPLEX in Section 6.3.1. In Section 6.3.2, the performance of the I-TS algorithm, greedy insertion algorithm, tabu search algorithm, and variable neighborhood search algorithm are compared by 50-request and 100-request instances.

**6.3.1 Results on the 20-request instances.** This subsection validates the correctness of the I-TS algorithm by solving six 20-request instances in non-peak periods. Let $\sigma^2$ ($V_r$) = 0, $\sigma^2$ ($V_f$) = 0. The results of the I-TS algorithm and CPLEX optimizer are compared. The maximum computational time of CPLEX optimizer is 30 minutes. The I-TS algorithm performs ten independent runs for each instance and computes the average value and best value. Table 3 summarizes the experimental results, where $F_{avg}$ is the average cost by the I-TS algorithm; $F_{bst}$ is the best objective value; $F_{CPLEX}$ is the objective value found by the CPLEX optimizer; and *Time*

**Table 2. Parameter settings.**

| Parameters | Value |
|---|---|
| The daily fixed cost of each AMR $\xi_1$ | 30$ |
| The penalty cost of violating time windows $\xi_2$ | 0.1$ |
| The unit travel cost of the AMR $\xi_1$ | 0.01$/$s$ |
| The variance of the AMR's traveling time $\sigma^2$ ($V_r$) | 0.15($s^2$/$m^2$) |
| The variance of elevator running time $\sigma^2$ ($V_f$) | 0.5($s^2$/$level^2$) |

https://doi.org/10.1371/journal.pone.0292002.t002

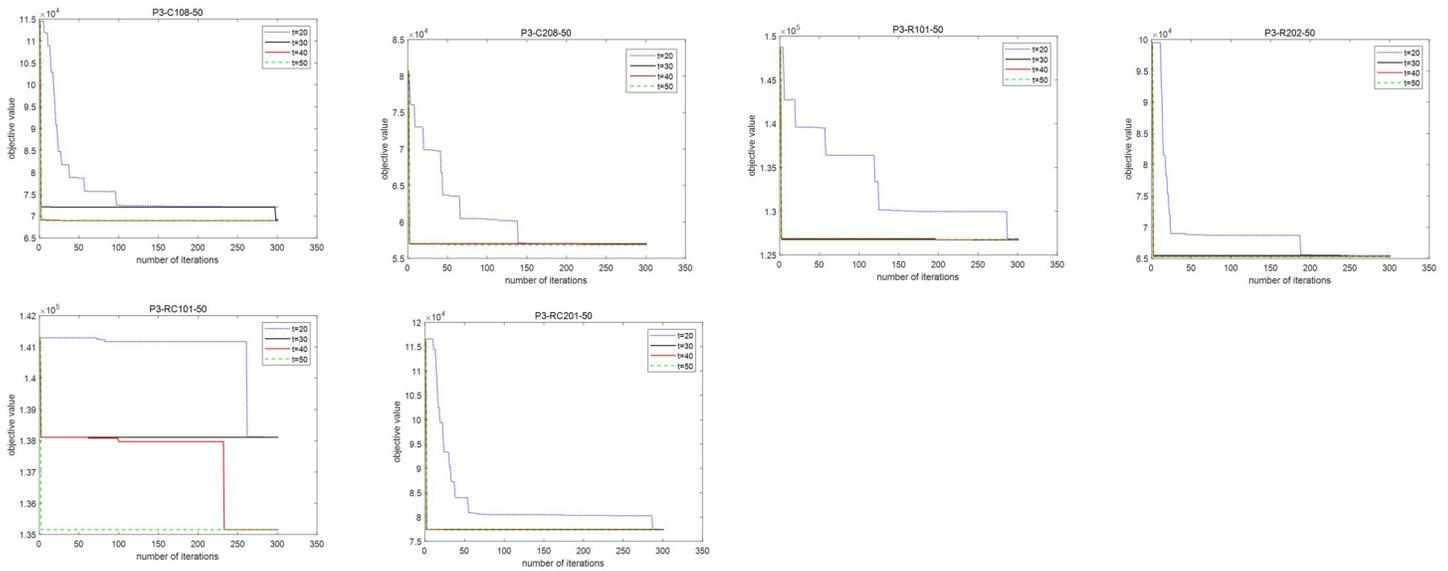

**Fig 7. Influence of different tabu tenure *t* on objective value.**

https://doi.org/10.1371/journal.pone.0292002.g007

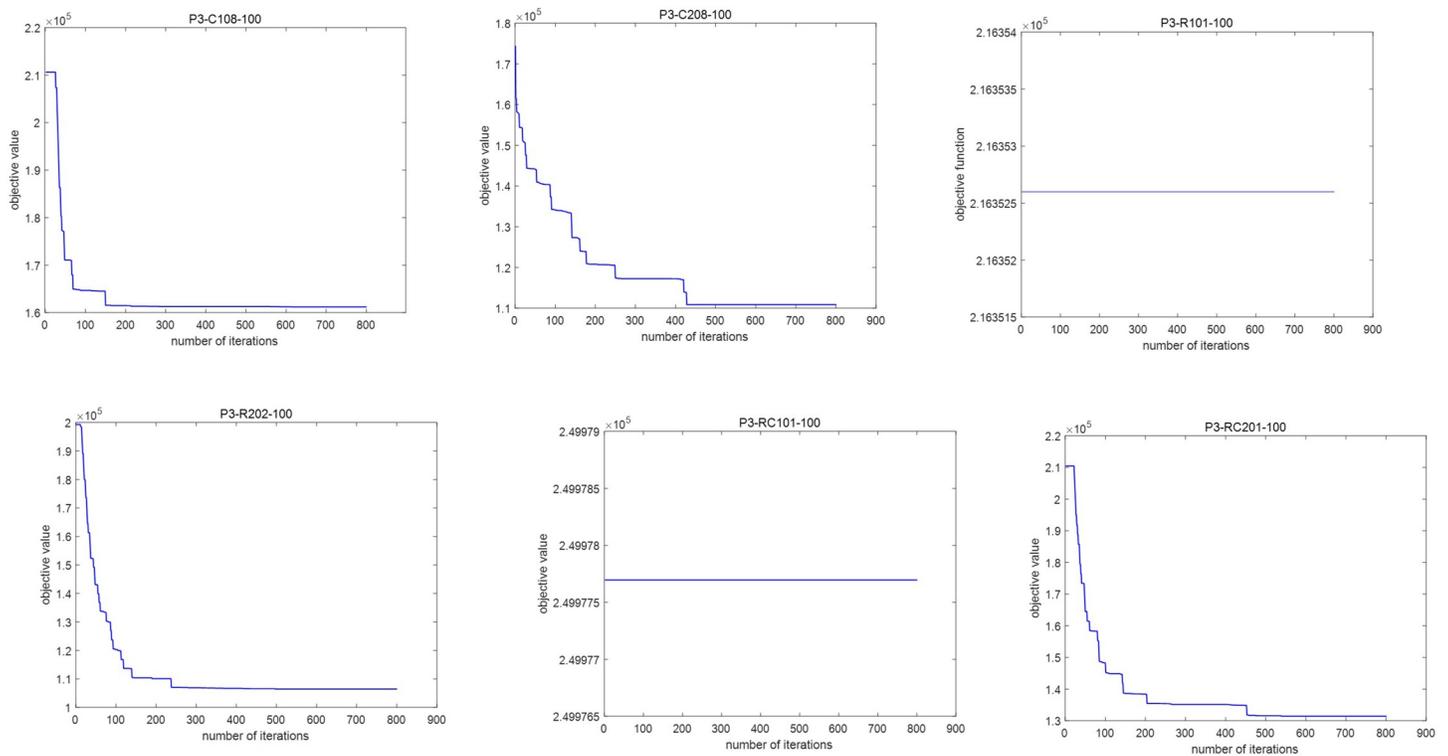

**Fig 8. Fitness value curve of total cost.**

https://doi.org/10.1371/journal.pone.0292002.g008

Table 3. Comparison of the I-TS algorithm and CPLEX.

| Instance | CPLEX | | I-TS | | | | |
|---|---|---|---|---|---|---|---|
| | $F_{CPLEX}$ | Time (s) | $F_{bst}$ | $G_1$ (%) | $F_{avg}$ | $G_2$ (%) | Time (s) |
| P3-C108-20 | **12823** | 118.1 | 13662.8 | 6.54 | 13742.69 | 7.17 | 6.14 |
| P3-C208-20 | **4071.5** | 37.58 | 4145.912 | 1.8 | 4151.103 | 1.95 | 7.29 |
| P3-R101-20 | **34616** | 303.8 | 34879.5 | 0.76 | 35211.1 | 1.71 | 7.01 |
| P3-R202-20 | 7100 | 1800* | **7067.309** | -0.46 | **7090.531** | -0.13 | 8.14 |
| P3-RC101-20 | 35193 | 1800* | **34971.31** | -0.62 | **35030.02** | -0.46 | 6.06 |
| P3-RC201-20 | **10179** | 33.49 | 10220.38 | 0.4 | 10240.28 | 0.6 | 7.73 |
| Average | **17330.4166** | 123.24 | 17491.2018 | **1.76** | 17577.6206 | **2.00** | 7.06 |

* Best feasible solution within 30 minutes.

https://doi.org/10.1371/journal.pone.0292002.t003

is the average runtime. Let

$$G_1 = \frac{F_{bst} - F_{CPLEX}}{F_{CPLEX}} \times 100\%; G_2 = \frac{F_{avg} - F_{CPLEX}}{F_{CPLEX}} \times 100\%.$$

The results indicate that CPLEX gets the optimal solution for only four instances (P3-C108-20, P3-C208-20, P3-R101-20, and P3-RC201-20) within 30 minutes. The optimal solution of the P3-R202-20 and P3-RC101-20 instances can't be found within 1800 seconds by CPLEX. But I-TS algorithm can find better solutions within 10 seconds. Table 3 shows that, the average error value of $|G_1|$ is 1.76%, and the average error value of $|G_2|$ is 2.00%. The difference of the optimal objective value by I-TS algorithm and CPLEX is small. But the I-TS algorithm can obtain good solutions in a relatively short time, and it even outperforms CPLEX in R202 and RC101 instances. This indicates the effectiveness of the I-TS algorithm in finding near-optimal solutions. The I-TS algorithm achieved competitive performance compared to CPLEX in terms of optimizing the objective function.

**6.3.2 Results on the 50-request and 100-request instances.** In this subsection, the effectiveness of the proposed I-TS algorithm is verified by the 50-request and 100-request instances in various environments. The performance of the I-TS algorithm, tabu search algorithm, greedy insertion algorithm, and variable neighborhood search (VNS) algorithm are compared.

The I-TS algorithm proposed in this paper is an enhancement of the greedy insertion algorithm and TS. In addition, the VNS algorithm is a widely used heuristic method for solving complex optimization problems, including combinatorial optimization problems such as the shortest circuit [41, 42]. We apply the greedy insertion algorithm, TS algorithm, and VNS algorithm to solve the AMR scheduling problem and compare the results with those obtained by the I-TS algorithm. For the TS algorithm and VNS algorithm, we randomly generate initial solutions. To ensure fairness, let the number of iterations ($N$ = 500) be equal for each algorithm. The results are presented in Table 4, where $F$ is the average cost found by the algorithm over ten runs, Time denotes the average runtime required for obtaining the solutions over ten runs, and $G'$ represents the percentage difference between the objective value achieved by a specific algorithm and that of the I-TS algorithm. Note that $G' > 0$ implies that the I-TS algorithm performs better.

The experimental results in Table 4 demonstrate that the I-TS algorithm significantly enhances the solution quality compared with the TS algorithm ($G'$ = 5.3044%). A higher-quality solution is generated by the I-TS algorithm in comparison to the TS algorithm, as it incorporates the greedy insertion algorithm for generating an initial solution. Furthermore, $G'$ = 0%

Table 4. Comparison of 50-request and 100-request instances in different environments.

| Instance | I-TS | | TS | | | greedy | | | VNS | | |
|---|---|---|---|---|---|---|---|---|---|---|---|
| | F | Time (s) | F | Time (s) | G' (%) | F | Time (s) | G' (%) | F | Time (s) | G' (%) |
| P1-C108-50 | **138131.2** | 32.58 | 142569.6 | 52.10 | 3.213 | **138131.2** | 0.218 | **0** | 146975.1 | 13.17 | 6.402 |
| P1-C208-50 | **138131.2** | 31.97 | 141407 | 52.13 | 2.371 | **138131.2** | 0.093 | **0** | 143990.1 | 13.32 | 4.241 |
| P1-R101-50 | **125659.8** | 33.00 | 131139.3 | 53.05 | 4.36 | 157194.9 | 0.031 | 25.09 | 136701.80 | 13.19 | 8.787 |
| P1-R202-50 | **116566** | 33.18 | 130258.4 | 53.12 | 11.74 | **116566** | 0.078 | **0** | 133465.3 | 14.36 | 14.49 |
| P1-RC101-50 | **142422.8** | 31.21 | 151726.1 | 60.38 | 6.532 | 156791.1 | 0.046 | 10.08 | 157109.1 | 12.22 | 10.31 |
| P1-RC201-50 | **115390.3** | 35.02 | 123253.1 | 54.6 | 6.814 | **115390.3** | 0.031 | **0** | 125836.6 | 15.53 | 9.052 |
| P1-C108-100 | **283649.2** | 166.7 | 283715.8 | 205.2 | 0.023 | 303108.6 | 0.218 | 6.860 | 290364.4 | 28.09 | 2.367 |
| P1-C208-100 | **119234.6** | 106.4 | 120038 | 215.9 | 0.673 | 190787.2 | 0.125 | 60.00 | 136410.2 | 30.98 | 14.40 |
| P1-R101-100 | **245287.5** | 103.4 | 282002.9 | 203.2 | 14.96 | **245287.5** | 0.093 | **0** | 296575.75 | 24.87 | 20.9 |
| P1-R202-100 | **252781.5** | 108.0 | 271933.5 | 207.0 | 7.576 | **252781.5** | 0.093 | **0** | 279382.5 | 28.80 | 10.52 |
| P1-RC101-100 | **260572.9** | 100.5 | 296428.8 | 221.4 | 13.76 | **260572.9** | 0.062 | **0** | 305556.9 | 25.42 | 17.26 |
| P1-RC201-100 | 249879.4 | 111.2 | **249257.2** | 210.7 | -0.24 | 277682.3 | 0.218 | 11.12 | 258361.1 | 31.01 | 3.394 |
| P2-C108-50 | **73986.29** | 30.55 | 74107.73 | 60.98 | 0.164 | 118706.9 | 0.109 | 60.44 | 76570.2 | 17.29 | 3.492 |
| P2-C208-50 | **43831.26** | 33.69 | 44164.2 | 63.22 | 0.759 | 81901.97 | 0.078 | 86.85 | 47472.89 | 19.14 | 8.308 |
| P2-R101-50 | **128011.7** | 36.59 | 130172.4 | 56.04 | 1.687 | 156474.9 | 0.046 | 22.23 | 136953.54 | 16.17 | 6.98 |
| P2-R202-50 | **52469.66** | 30.98 | 62463.19 | 60.77 | 19.04 | 97529.78 | 0.078 | 85.87 | 56293.26 | 18.29 | 7.287 |
| P2-RC101-50 | **139689** | 46.3 | 141537.7 | 94.81 | 1.323 | 146172.9 | 0.109 | 4.641 | 144563.7 | 12.78 | 3.489 |
| P2-RC201-50 | 62516.33 | 38.4 | **62463.19** | 76.34 | -0.08 | 118200.5 | 0.078 | 89.07 | 66876.11 | 17.78 | 6.973 |
| P2-C108-100 | **164004.3** | 138.8 | 164215.6 | 239.8 | 0.128 | 213082.1 | 0.093 | 29.92 | 171947.9 | 34.82 | 4.843 |
| P2-C208-100 | **117677.2** | 145.2 | 118068.6 | 245.4 | 0.332 | 183445.4 | 0.078 | 55.88 | 137075.4 | 35.64 | 16.48 |
| P2-R101-100 | **241042.2** | 113.8 | 280012.4 | 210.7 | 16.16 | **241042.2** | 0.171 | **0** | 293338.31 | 30.69 | 21.69 |
| P2-R202-100 | **112837.4** | 199.4 | 112947.7 | 248.6 | 0.097 | 201592 | 0.031 | 78.65 | 131041.2 | 37.30 | 16.13 |
| P2-RC101-100 | **252827.8** | 194.0 | 278140.1 | 299.4 | 10.01 | **252827.8** | 0.125 | **0** | 289855.6 | 26.83 | 14.64 |
| P2-RC201-100 | 135960.1 | 137.7 | **135524** | 261.5 | -0.32 | 234969.2 | 0.156 | 72.82 | 154032.8 | 35.95 | 13.29 |
| P3-C108-50 | **71421.05** | 35.59 | 72053.17 | 51.75 | 0.885 | 117574.8 | 0.031 | 64.62 | 74834.39 | 17.08 | 4.779 |
| P3-C208-50 | **42017.33** | 37.20 | 42598.03 | 158.31 | 1.382 | 80741.28 | 0.046 | 92.16 | 45638.3 | 19.14 | 8.617 |
| P3-R101-50 | **126886.3** | 30.80 | **126886.3** | 48.78 | 0 | 148727.4 | 0.031 | 17.21 | 131256.18 | 16.30 | 3.44 |
| P3-R202-50 | **49802.75** | 31.40 | 50892.49 | 115.58 | 2.188 | 99498.61 | 0.031 | 99.78 | 55330.96 | 18.42 | 11.10 |
| P3-RC101-50 | **136872.9** | 36.31 | 138383.5 | 119.62 | 1.103 | 141288.6 | 0.062 | 3.226 | 141624.6 | 12.74 | 3.471 |
| P3-RC201-50 | **60680.25** | 33.61 | 61482.91 | 134.26 | 1.322 | 113628 | 0.062 | 87.25 | 64762.52 | 18.23 | 6.727 |
| P3-C108-100 | **160349.2** | 147.2 | 162159.5 | 221.29 | 1.129 | 210586.6 | 0.093 | 31.33 | 169171.3 | 35.89 | 5.501 |
| P3-C208-100 | **114986.1** | 172.5 | 115567.6 | 232.95 | 0.505 | 180605.8 | 0.062 | 57.06 | 137379.4 | 36.0 | 19.47 |
| P3-R101-100 | **216400** | 110.5 | 272455.4 | 195.85 | 25.9 | 225915.7 | 0.046 | 4.397 | 285628.67 | 31.87 | 31.99 |
| P3-R202-100 | **110553.3** | 160.1 | 131192.7 | 279.69 | 18.66 | 199281.6 | 0.078 | 80.25 | 149853.3 | 37.41 | 35.54 |
| P3-RC101-100 | **249977.7** | 117.4 | 289225.5 | 259.60 | 15.70 | **249977.7** | 0.062 | **0** | 295845.5 | 27.11 | 18.34 |
| P3-RC201-100 | **131633.7** | 151.1 | 133102.4 | 214.21 | 1.115 | 229341.4 | 0.062 | 74.22 | 151737.9 | 39.89 | 15.27 |
| Average | 144003.5 | **86.17** | 153431.8 | 153.83 | 5.304 | 177653.8 | 0.086 | 36.41 | 161661.46 | 23.99 | 11.38 |

https://doi.org/10.1371/journal.pone.0292002.t004

for the greedy insertion algorithm indicates that the initial solution generated by the greedy insertion algorithm is of high quality and can greatly aid in finding better solutions. Similarly, the optimal solution obtained by the classical VNS algorithm using the same number of iterations is inferior to that achieved by the I-TS algorithm (by 11.38%). Although the I-TS algorithm takes longer to obtain the optimal solution, its computational time is acceptable considering the quality of the generated solution. When strict computation time requirements exist, the greedy insertion algorithm and VNS approach can be used to solve the problem, since their computational time is shorter than that of the I-TS algorithm.

To further illustrate the performance of the I-TS algorithm, the fitness curves for the three algorithms (I-TS, TS, and VNS) in non-peak cases are depicted in Fig 9. It shows that the I-TS algorithm achieves less objective value compared to the other algorithms while the number of iterations is large enough (bigger than 100).

### 6.4 AMR routing planning

In this section, we take an example to analyze the specific service routes of AMRs. Consider the RC101-20 instance in three periods (morning peak, afternoon peak, and non-peak). The AMR route planning of instance RC101-20 is shown in Table 5, where *No.* is the AMR sequence number, and *Route* is the running route of the AMR (0 represents the depot, 1–20 represents the medical request of instance P1-RC101-20, 21–40 represents the medical request of P2-RC101-20 instance, and 41–60 represents the medical request of P3-RC101-20 instance). *AT* is the time when the AMR arrives at the request, and *r* is the service probability of the route.

The results presented in Table 5 reveal the following findings: (1) In order to effectively handle the tasks, a minimum of 16 AMRs is required, with each AMR making at least two trips. (2) Half of the AMRs achieve a service probability of 1 on their planned routes. (3) As the number of requests on the route increases, the service probability tends to decrease. The minimum service probability can reach 0.97. These outcomes underscore the effectiveness of the proposed stochastic programming model in planning the number of AMRs and service routes across diverse environments.

### 6.5 Management insight

In order to demonstrate the service quality of AMRs and validate the stochastic programming model, we introduce the concept of the service probability (*r*), which represents the probability of medical requests being serviced within a specified time window. We compare the *r* values of 50-request and 100-request instances in different environments. Eq (12) defines the service probability *r*.

$$r = \min_i P\left\{A_{pi}^k \leq h_i\right\} \tag{12}$$

Each instance is run ten times, and the average service probability $r_{avg}$ is given in Table 6.

According to Table 6, the service probability during the morning peak (represented by the red column) is less than or equal to that in the afternoon peak (represented by the green column) and non-peak hours (represented by the blue column). The average service probability is over 90% in all environments except for C208-100 in the morning peak. Even during the afternoon peak, the probability of medical requests receiving services within a specified time window can reach 0.99. These results are relatively satisfactory for route planning problems of AMRs with high randomness.

Fig 10 compares of the number of AMRs participating in service for different instances. It shows that in the same environment, the C1, R1, and RC1 datasets with narrower time windows require a greater number of AMRs than the C2, R2, and RC2 datasets with wider time windows. This difference arises because wider time windows have a lower likelihood of time window violations, leading to decreased penalty costs. During the morning peak period, the environment is more intricate, leading to heightened obstacles and extended travel durations for the AMRs. Consequently, to ensure the timely fulfillment of medical requests within their respective time windows, an increased number of AMRs becomes imperative.

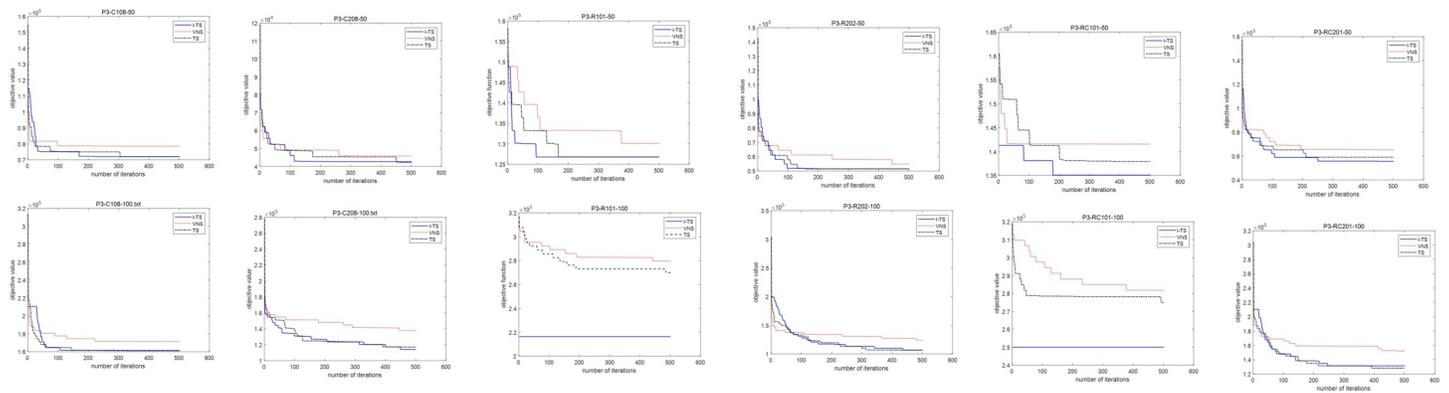

**Fig 9. Fitness curve of total cost.** (a) 50-request instances. (b) 100-request instances.

https://doi.org/10.1371/journal.pone.0292002.g009

The proposed stochastic programming model is feasible for the stochastic scheduling of AMRs in different environments at hospitals. Based on the environment of medical requests, we can allocate the appropriate number of AMRs to complete the service, thereby minimizing the hospital's daily cost and maximizing the AMR utilization rate.

## 7 Conclusion

This paper investigates the scheduling problem of AMRs at hospitals, where service time and AMR travel time are random. We establish a stochastic programming model for the AMR scheduling problem, which minimizes the sum of fixed cost, AMR operation cost, and delayed service cost. To solve the problem, we propose a heuristic algorithm that combines the greedy insertion algorithm and tabu search algorithm. A repair operator is introduced to transform infeasible solutions into feasible ones. The experimental results demonstrate that the I-TS algorithm can generate reliable service routes for AMRs in stochastic environments. Moreover, by

**Table 5. AMR routing planning of instance RC101-20.**

| No. | Route | AT | r |
|---|---|---|---|
| 1 | 0-14-0-35-32-33-0-47-41-0 | 08:00:35–08:01:34–12:00:58–12:01:16–12:02:07–12:06:39–14:01:05–14:01:55–14:05:58 | 0.98 |
| 2 | 0-5-1-0-25-27-21-0-50-57-0 | 08:00:41–08:00:58–08:10:11–12:00:41–12:01:03–12:01:51–12:05:31–14:01:59–14:02:40–14:04:32 | 0.97 |
| 3 | 0-2-0-26-23-0-46-43-0 | 08:00:50–08:02:36–12:01:35–12:01:56–12:03:49–14:01:35–14:01:59–14:04:07 | 0.99 |
| 4 | 0-15-13-0-40-0-52-53-0 | 08:00:58–08:01:55–08:07:31–12:02:02–12:04:02–14:01:04–14:01:56–14:04:36 | 0.99 |
| 5 | 0-11-0-31-0-58-60-0 | 08:00:59–08:02:57–12:00:59–12:01:45–14:01:27–14:01:50–14:04:29 | 1 |
| 6 | 0-12-17-0-38-0-54-0 | 08:01:04–08:01:40–08:06:34–12:01:27–12:03:36–14:00:35–14:01:30 | 0.99 |
| 7 | 0-16-0-24-0-56-0 | 08:01:12–08:02:40–12:02:21–12:03:58–14:01:12–14:02:35 | 1 |
| 8 | 0-19-0-22-0-44-0 | 08:01:12–08:03:50–12:00:50–12:01:33–14:02:21–14:04:14 | 1 |
| 9 | 0-7-0-29-0-49-0 | 08:01:19–08:03:07–12:01:31–12:02:59–14:01:31–14:03:13 | 1 |
| 10 | 0-18-0-39-0-42-0 | 08:01:27–08:04:09–12:01:12–12:03:19–14:00:50–14:01:40 | 1 |
| 11 | 0-8-0-30-0-55-0 | 08:01:31–08:03:38–12:01:59–12:03:17–14:00:58–14:02:30 | 1 |
| 12 | 0-9-0-48-0 | 08:01:31–08:03:20–14:01:31–14:03:29 | 1 |
| 13 | 0-6-4-0-59-0 | 08:01:35–08:01:52–08:09:06–14:01:12–14:03:39 | 0.98 |
| 14 | 0-3-0-51-0 | 08:01:49–08:03:57–14:00:59–14:01:52 | 1 |
| 15 | 0-10-0-34-36-0 | 08:01:59–08:06:25–12:00:35–12:01:12–12:02:54 | 0.99 |
| 16 | 0-20-0-28-37-0 | 08:02:02–08:04:32–12:01:31–12:02:26–12:05:15 | 0.98 |

https://doi.org/10.1371/journal.pone.0292002.t005

Table 6. Average service probability $r_{avg}$ of three periods.

| $r_{avg}$ | C108-50 | C208-50 | R101-50 | R202-50 | RC101-50 | RC201-50 | C108-100 | C208-100 | R101-100 | R202-100 | RC101-100 | RC201-100 | Average |
|---|---|---|---|---|---|---|---|---|---|---|---|---|---|
| Morning peak | 0.98 | 0.98 | 0.97 | 0.94 | 0.99 | 0.90 | 0.97 | 0.79 | 0.98 | 0.94 | 0.98 | 0.92 | 0.945 |
| Afternoon peak | 0.99 | 0.98 | 0.97 | 0.99 | 0.99 | 0.99 | 0.99 | 0.94 | 0.97 | 0.99 | 0.97 | 0.99 | 0.98 |
| Non-peak | 0.99 | 0.98 | 0.98 | 0.99 | 0.99 | 0.99 | 0.99 | 0.95 | 0.98 | 0.99 | 0.98 | 0.99 | 0.98 |

https://doi.org/10.1371/journal.pone.0292002.t006

allowing each AMR to handle multiple trips, the hospital's overall expenses can be minimized and the AMRs' efficiency can be increased.

In this study, we assume that both the AMR's running time and service time follow normal distributions, and all requests are known before planning the route. In the future, we can

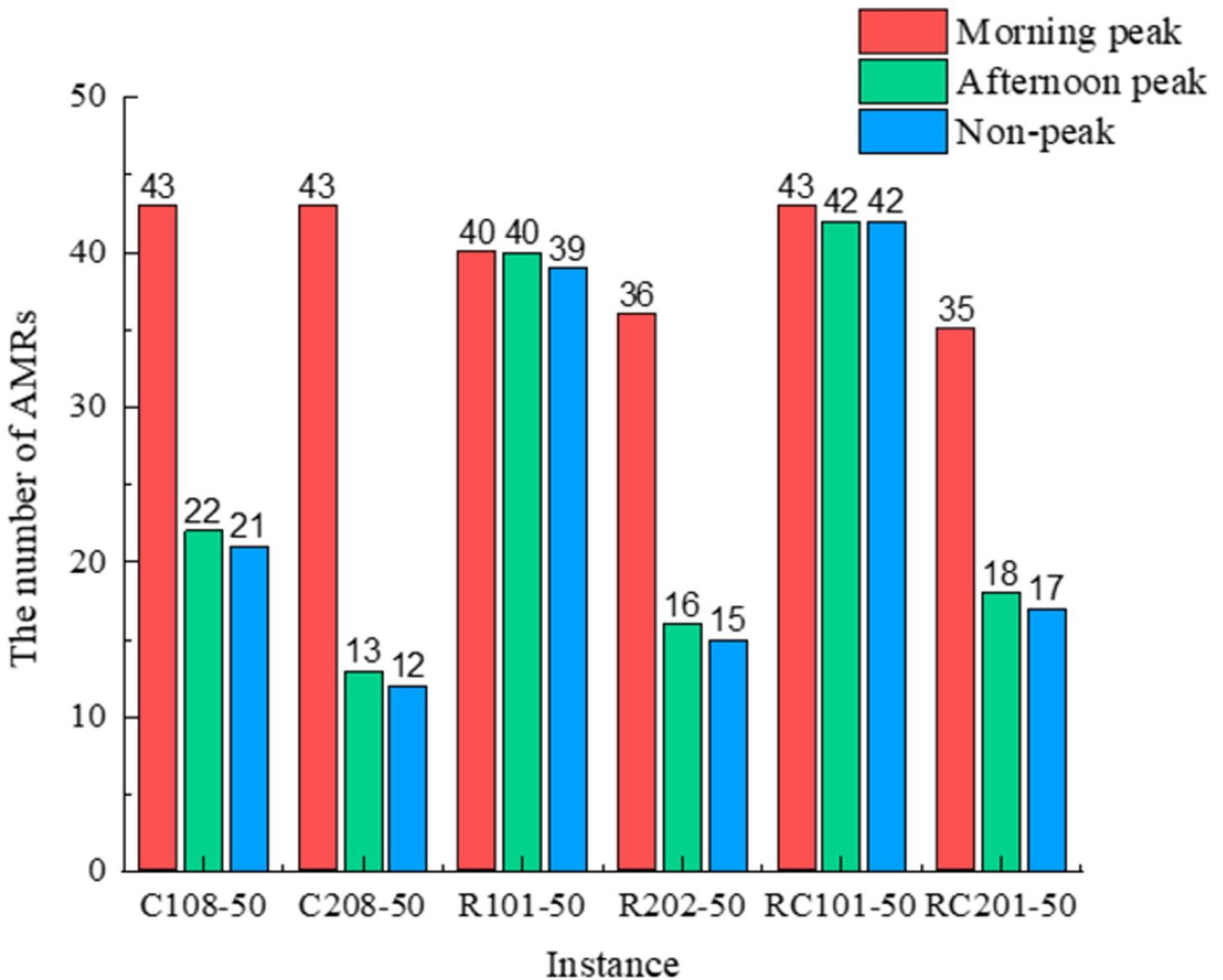

Fig 10. Number of AMRs in different environments.

https://doi.org/10.1371/journal.pone.0292002.g010

explore the AMR scheduling problem with dynamic emergency requests and more general distributed running and service times, which would further complicate the problem.

## Supporting information

**S1 File. New instances based on three instances in the Solomon benchmark (C, R, and RC).**
(ZIP)

## Author Contributions

**Conceptualization:** Ning Zhao.

**Data curation:** Lulu Cheng.

**Formal analysis:** Lulu Cheng.

**Methodology:** Lulu Cheng, Ning Zhao, Mengge Yuan, Kan Wu.

**Software:** Lulu Cheng.

**Validation:** Lulu Cheng.

**Visualization:** Lulu Cheng.

**Writing – original draft:** Lulu Cheng.

**Writing – review & editing:** Lulu Cheng, Ning Zhao.